CHAPTER 2

# Benchmarking Sim2Real Gap: High-fidelity Digital Twinning of Agile Manufacturing

S. Katyara[a], S. Sharma[b], P. Damacharla[c], C. Garcia-Santiago[a], L. Dhirani[d], B. S. Chowdhry[e]

[a]IMR Ltd, Ireland; [b]Shutterfly INC., USA; [c]KineticAI INC., USA; [d]University of Limerick, Ireland; [e]Mehran UET, Jamshoro

ABSTRACT
As manufacturing industry shifts from mass production to mass customization, there is a growing emphasis on adopting agile, resilient, and human-centric methodologies that are in line with the directives of Industry 5.0. The key to this transformation is the deployment of digital twins technology that digitally replicates manufacturing assets to enable enhanced process optimization, predictive maintenance, generation of synthetic data, and accelerated customization and prototyping. This chapter delves into the technologies underpinning the creation of digital twins specifically designed for agile manufacturing scenarios within the domain of robotic automation. We explore the idea of transfer of trained policies and process optimizations from simulated settings to real-world applications by exploiting techniques including domain randomization, domain adaptation, curriculum learning, and model-based system identification. The chapter also examines various industrial manufacturing automation scenarios such as; bin-picking, part inspection, and product assembly within Sim2Real conditions. The performance of digital twin technologies in these scenarios is evaluated using practical metrics i.e., data latency, adaptation rate, simulation fidelity among others, thus providing a comprehensive assessment of their efficacy, effectiveness and potential impact on modern manufacturing processes.

## 1. Introduction

As the manufacturing industry advances towards the vision of Industry 5.0, the digital twins are emerging as a transformative technology that redefine the possibilities and methodologies across the manufacturing landscape and beyond. Prominent organizations including NASA, Tesla and NVIDIA all are exploiting digital twin technologies under diverse applications such as product development, quality control, predictive maintenance and asset management (1). A recent report (2) predicts that the global market for digital twins will reach to $155.84 billion by 2030 with a compound annual growth rate (CAGR) of 35.7% from 2024 to 2030. This rapid growth highlights a significant shift towards the agile manufacturing practices which are essential for adapting to not only increasing demands of mass customization and responding to market volatility but also integrating continuous technological advancements. The convergence of artificial intelligence (AI), the Internet of Things (IoT), today's robotics, and data analytics within the digital twin frameworks enhances their precision and functionality hence enabling sophisticated end-to-end bi-directional connectivity and data-driven decision-making. This integration is crucial for developing resilient manufacturing systems that can not only predict disruptions and optimize product development in real-time but also efficiently manage production of customized orders at large scale.

CONTACT: S. Katyara Email: sunny.katyara@imr.ie

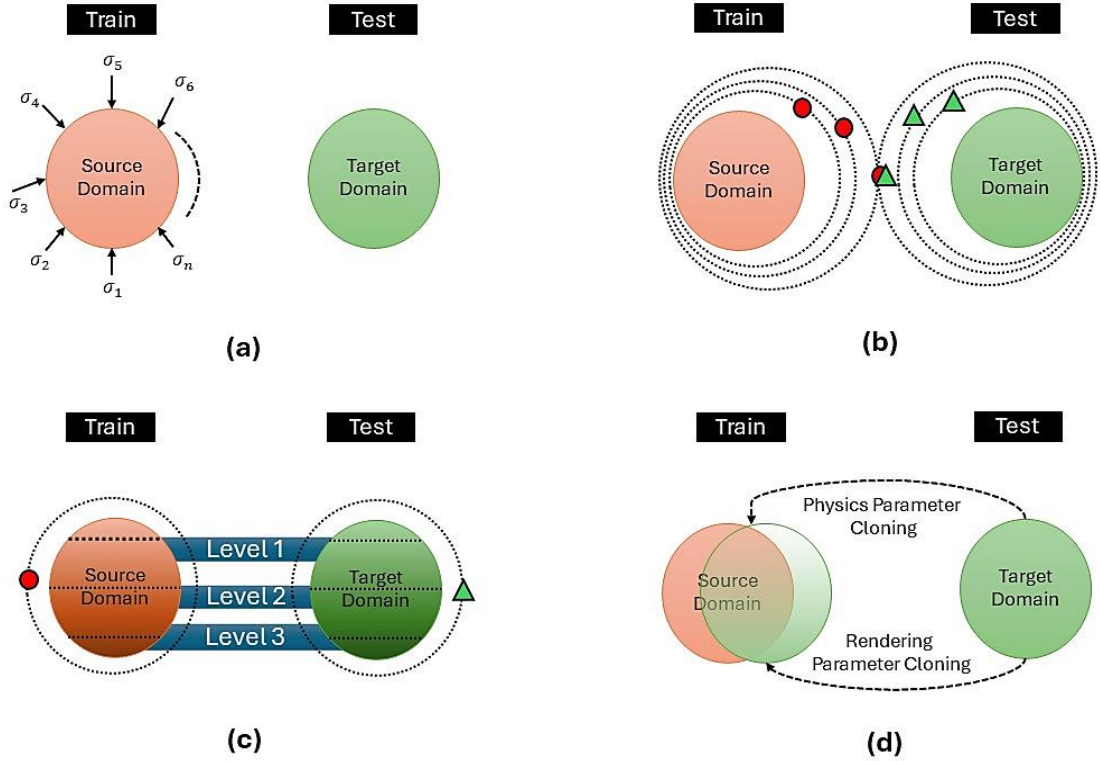

**Figure 1.** Sim2real transfer approaches for agile manufacturing automation; (a) represents Domain Randomization (b) shows Domain Adaptation (c) illustrates Curriculum learning (d) depicts Model-based System Identification.

The implementation and choice of digital twin technologies differ across industries, as being tailored to meet the specific operational requirements and strategic goals. Siemens for instance employs its MindSphere platform (3) which integrates IoT data within the digital twins to deliver actionable insights into its operations, maintenance and the optimization processes. Likewise, the BMW has partnered with NVIDIA to harness their advanced Omniverse platform for creating a digital twin of their manufacturing facility to minimize errors, improve production flows and also to enhance overall efficiency (4). General Electric (GE) on the other hand utilizes its custom cloud-based platform so called Predix (5) that is not only used for real-time monitoring of equipment health and failure prediction but also for proactive maintenance scheduling. These are the strategies that reduces downtime and also extend the operational lifespan of company's products. Such examples among many others in the manufacturing sector underscore the growing adoption of digital twins by leading companies to enhance their effectiveness, efficiency, resilience and responsiveness against unforeseen conditions.

However, also the definition and interpretation of digital twinning vary significantly across distinct industries. For some, it represents a simulated setup (6) while for others it serves as a virtual environment for prototyping, product design and process optimization (7) and yet for others, it is a tool for visualization and monitoring (8). This chapter seeks to consolidate these perspectives and propose a unified definition of digital twinning. For us, the *digital twinning is defined as a closed-loop, dynamic virtual representation of physical assets, whether processes, objects, or entire systems, throughout their lifecycle thereby leveraging real-time data to accurately reflect the state, structure and behavior of the entity being virtualized*. It is important to distinguish between the terms digital twin and simulations which are often being used interchangeably. Specifically, simulations, such as those used in robotics are specialized form of digital twins that are characterized by offline, non-dynamic updates and are typically designed under near-ideal conditions for solution design, development and evaluation. The inherent limitations of robotic simulations, particularly in their inability to be updated in real-time to reflect changes in their physical counterparts are comprehensively explored in detail in (9).

To effectively deploy digital twins in robotic automation scenarios, it is essential to bridge the gap between the simulated environments and real-world conditions. This process is known as Sim2Real transfer. Adapting digital twins o real-world applications requires overcoming the inherent differences between the virtual settings and the complexities of the physical environments. Such an adaptation is crucial for ensuring that the robots can perform reliably when faced with the unpredictable and dynamic nature of the real-world conditions. In order to achieve this, it requires the application of advanced Sim2Real transfer techniques including domain randomization, domain adaptation, curriculum learning and model-based system identification. The **Domain Randomization** as depicted in **Figure 1(a)** introduces distinct controlled random variations into the simulated environment such as; changing lighting, textures, geometries, and physical dynamics. This approach trains the robots to cope with unpredictability and generalize their skills to real-world conditions. By exposing the robot model to such variations during training phase, it prevents the model from over-fitting to the idealized conditions of simulation hence it improves its real-world robustness. In contrast, the **Domain Adaptation** focuses on refining the algorithms that govern the digital twin to ensure effective operation in dynamic and unpredictable environments. As shown in **Figure 1(b),** the domain adaptation techniques are critical for closing the fidelity gap between the simulation and reality. These techniques adjust the robot's control policies and sensor models to compensate for any discrepancies in the sensor readings, physical properties and interaction dynamics encountered in the real world situations, hence ensuring that the digital twin performs reliably stable across different task contexts. However the **Curriculum Learning**, as shown in **Figure 1(c)**, involves training robots through a sequence of progressively more complex tasks similar to how human learning process works. This method begins with simpler, low-variance scenarios and gradually introducing more complexity thus allowing the robot to build on its previous knowledge systematically. This stage-wise learning process stabilizes the robot's learning curve by reducing the risk of catastrophic forgetting and enhances its ability to tackle complex real-world tasks with greater resilience. Moreover, the **Model-based System Identification** as shown in **Figure 1(d)**, leverages physics-informed and domain-specific models to predict the future states of the environment during learning process. This technique enables the robot to plan its actions more effectively by considering the long-term consequences of its decisions. Therefore, by integrating accurate simulations with real-world data, the model-based system identification provides a safe and efficient framework for robots to explore and optimize their decision variables that would be risky or infeasible to test directly onto the physical setup. All these Sim2Real strategies are especially crucial for scenarios that demand real-time adaptability and decision-making based on the long-term outcomes. By incorporating such advanced techniques, the digital twins can bridge the fidelity-gap between the simulation and reality thus ensuring that the robots perform reliably and efficiently under complex and dynamic conditions.

This chapter delves into the array of technologies that have been utilized in the digital twinning of agile manufacturing processes and robotic automations. Additionally, it introduces innovative methods for creating digital twins exploited by IMR, as depicted in Figure 3 and further elaborated in Section II. The chapter also details the implementation of methods for Sim2Real transfer thus highlighting how they facilitate the efficient transfer of trained policies and algorithms for enabling faster and optimized adaptation in the real-world scenarios that being elaborated in Section III. Furthermore, a comprehensive benchmarking study is conducted to evaluate the effectiveness of these digital twin technologies using key practical metrics including success rate, localization accuracy, cycle time, commissioning time, reprogramming time, flexibility rate, repeatability rate, among others. This study assesses three distinct scenarios i.e., bin-picking, part inspection and product assembly tasks, that provide in-depth analysis and discussion in Section IV. The chapter concludes with key insights and recommendations of using advancing digital twin technologies and Sim2Real transfer strategies. These conclusions with major focus on industry applications aim to maximize the benefits of such technologies and offer a roadmap for future developments, ensuring that the digital twins continue to enhance efficiency, adaptability and innovation into the agile manufacturing industry.

2. Digital Twin Technologies

When selecting digital twin technology from commercially available options for robotic automation in manufacturing, several critical factors must be carefully evaluated, including domain feasibility, task complexity, computational resources and ease of use (10). The quality of simulations is of paramount importance with a particular focus on the accuracy and realism of the virtual environment as they directly impact the effectiveness of Sim2Real transfer efforts. The data integration capabilities are also crucial for seamless integration with existing data sources and IoT devices that enables real-time monitoring and feedback loops, that are essential for dynamic operations. The advanced analytics and AI-driven insights further enhance the decision-making process by not only providing predictive maintenance and optimization opportunities but also the process improvements advantages. In addition to technical capabilities, the accessibility and ease of implementation of digital twin technology need to be well aligned with the expertise of in-house team and the available resources. Scalability is another key consideration, to ensure that the chosen digital twin solution can accommodate the future growth and adapt to evolving manufacturing needs without requiring significant overhauls. Maintaining a balance between the cost and return on investment is critical under especially high-stakes manufacturing environments where the efficiency and profitability are closely tied. It is also important to select digital twin technology that has strong community support and receives regular updates ensuring continuous improvement and compatibility with the emerging technologies. Security and data privacy are non-negotiable, requiring adherence to the industry standards and ensuring that the solution is interoperable with existing systems. Additionally, compatibility with other enterprise systems, such as MES, ERP and PLM, is important for creating a cohesive digital ecosystem. Based on levels of fidelity and functionality, the digital twins for robotic automation, as depicted in **Figure 2**, can be characterized into different categories, each suited to the specific use cases and operational requirements. These categories include geometric and physics-based twins for basic monitoring and control, data-drive twins for enhanced process optimization and the hybrid and AI-enhanced twins for complex simulations and real-time decision-making.

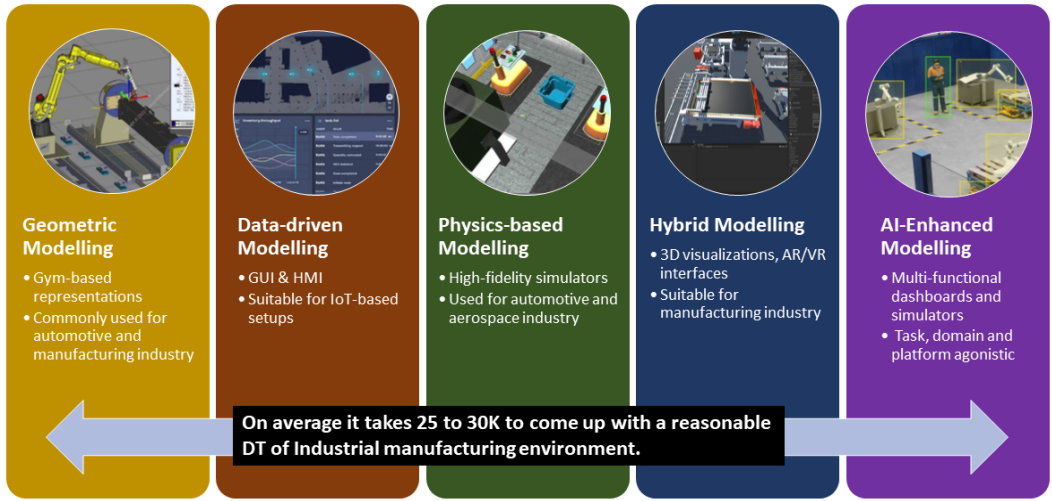

Figure 2. Digital twin technologies for agile robotic automation

- **Geometric Modelling**: Geometric digital twins are 3D virtual representations of the robots and their auxiliary work cells that are created using CAD models or 3D scans of parts. This technology is primarily employed for visualization, layout optimization, reachability analysis, virtual commissioning and operator training. However, the geometric digital twins typically lack real-time data integration and dynamic behaviour representation that limit their applicability in scenarios requiring predictive maintenance or real-time control. Despite these limitations, the geometric digital twins provide valuable insights for planning and design. For more comprehensive analysis and optimization of robotic automation processes, more advanced digital twins such as data-driven or hybrid models are rather preferred, as they offer more dynamic understanding of robotic systems and facilitate better decision-making and operational efficiency.
- **Data-driven Modelling:** The data-driven digital twins for agile manufacturing automation leverage real-time sensory data from the physical robotic assets to create the live models of system that mirrors the actual state and behaviour of the physical setup. These digital twins support real-time monitoring and control that are used for predictive maintenance through advanced data analytics and enhance performance optimization by comparing simulated outcomes with the real-world data. The data-driven digital twins are particularly effective in anomaly detection for identifying deviations from expected operations that could indicate emerging issues. However, the success of data-driven digital twins is not only contingent upon the quality and integrity of data but also the robustness of the supporting infrastructure and the stringent security measures. Despite these challenges, the data-driven digital twins offer substantial benefits, including increased operational efficiency, reduced downtime and improved manufacturing processes.
- **Physics-based Modelling:** Physics-based digital twins extend beyond the simple visualization by simulating the physical interactions and the behaviours of robots and their systems under different operating conditions. By incorporating sophisticated models such as computational fluid dynamics (CFD) and finite element analysis (FEA), such digital twins enable engineers to optimize their robotic designs, conduct virtual testing, validate solutions at early development stages, perform what-if analyses and simulate complex robotic systems. This approach leads to significant improvements into the performance, efficiency and reliability while reducing development time and costs. However, designing and utilizing physics-based digital twins is computationally intensive thus require specialized expertise and powerful hardware. Additionally, the accuracy of these models and their successful validation and calibration are critical for effective Sim2Real transfer instances.

- **Hybrid Modelling:** Hybrid digital twins for agile manufacturing automation aim to seamlessly integrate geometric models (from both 3D scanning and CAD models), the real-time sensor data (i.e., vision, force, and position sensors), and the physics-based simulations (like CFD and FEA) to create a holistic virtual representation of the robotic systems under consideration. This integration enables comprehensive simulations of entire production lines, including robotic movements, interactions with objects, and the behaviour of other equipment. The hybrid digital twins are instrumental in virtual commissioning, predictive maintenance through advanced anomaly detection, and the real-time process optimization, where the robotic parameters are adjusted based on the live sensory feedback. By incorporating machine learning and artificial intelligence, the hybrid digital twins provide more advanced analytics and decision support, enabling process optimization and autonomous operations. However, the complexity of integrating these diverse elements requires specialized expertise in the system integration and data validation and the computational demands can be substantial and instrumental. Despite these challenges, the enhanced efficiency, productivity and informed decision-making offered by hybrid digital twins make them a valuable asset for Industry 4.0 applications.

- **AI-enhanced Modelling:** AI-enhanced digital twins, as illustrated in **Figure 3** employ cutting-edge AI technologies including deep learning, reinforcement learning and the computer vision models to analyse data streams from various digital twin models (i.e., geometric, physics-based, and data-driven). This enables real-time process optimization and advanced anomaly detection through pattern recognition in sensory data that allows for proactive maintenance by predicting the potential failures. By processing large volumes of historical and real-time data, the AI models can identify bottlenecks and inefficiencies thus optimizing the process parameters, robotic control actions and enabling autonomous decision-making for real-time adjustments. These digital twins also provide decision support to human operators by offering insights and recommendations based on the thorough data analyses. However, the effectiveness of AI-enhanced digital twins depends upon access to large and high-quality datasets for training and the availability of specialized AI expertise for model development and interpretation hence presenting significant challenges.

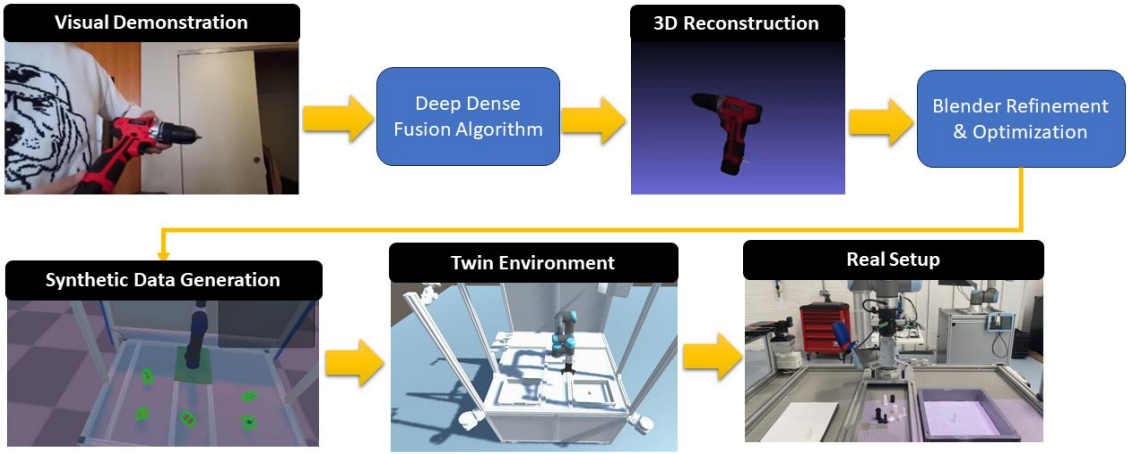

**Figure 3.** End-end AI-enhanced high-fidelity digital modeling and model2real transfer for agile manufacturing scenario under stochastic task conditions.

3. Sim2Real Technologies

Sim2Real transfer is essential for bridging the gap between the high-fidelity simulations and the real-world robotic automation setups. By selecting an appropriate Sim2Real transfer technique for agile robotic automation requires to take into consideration several critical factors including; computational power, task complexity, simulator fidelity and the specific requirements of the application (11) (12). For relatively simpler tasks such as he bin-picking, the techniques like domain randomization may be sufficient to ensure the robust performance by exposing the system to a wide variety of simulated parametric variations. However for more complex tasks, such as precision inspection or the intricate parts assembly often require more advanced approaches like domain adaptation, curriculum learning or the system identification to achieve the necessary level of accuracy and reliability. Additionally, the availability of high-quality real-world data and substantial computational resources play an important role in the success of Sim2Real techniques. For instance, the data-driven modelling may necessitate extensive datasets for training and validation as well as the robust infrastructure to handle real-time data integration and processing. Similarly, the physics-based models may require significant computational power to simulate complex interactions accurately. Ultimately, it is crucial to carefully select a digital twin approach that aligns well with the desired level of performance while ensuring balance between the complexity of the task and the available resources. This involves not only choosing the right Sim2Real method but also ensuring that the digital twin can support the necessary fidelity, scalability and the adaptability, that is desired for the specific robotic automation application. By making informed decisions in this selection process, the manufacturers can optimize their systems for efficient, reliable and the scalable operations in real-world environments. The following descriptions provide additional insights into the previously discussed Sim2Real transfer strategies. This is specifically tailored towards enhancing the understanding from the perspective of agile manufacturing automation rather than the generic concepts. They include;

- **System Identification**: The system identification is a reliable technique for achieving accurate Sim2Real policy transfer by calibrating simulations using the real-world data samples. This process involves collecting data from various robotic sensors including joint encoders, cameras and the force/torque sensors during the task execution such that to capture the robot's dynamics accurately. The choice of modelling approach whether it be linear (state-space models or transfer functions) or nonlinear (neural networks or Gaussian processes) is also important and that depends upon balancing model complexity against the accuracy. The parameter estimation techniques including least squares, maximum likelihood and the subspace methods are then employed to fit the model to the collected data. The validation phase compares the predictions of the models against new, unseen real-world data to ensure that the high fidelity is achieved and ensured. Once validated, the refined model is then integrated into the simulation environment either by replacing or augmenting the existing models. This integration ensures that the robot's behaviour in simulation closely mirrors its real-world counterpart thus allowing for more precise development and testing of control policies thereby facilitating reliable Sim2Real policy transfer without compromising the performance.

- **Domain Randomization**: The domain randomization is a robust Sim2Real transfer method that enhances the generalization capability of control policies by exposing the simulated environment to a wide variety of parametric variations. Such variations include visual aspects such as object shapes, sizes, textures, lighting and camera properties and as well as the physical attributes like friction, mass, contact dynamics and gravity. Additionally, the sensor parameters, including noise levels, delays and the potential failures are thus randomized during training phase. This extensive variability ensures that the trained model can handle the broad spectrum of conditions encountered in the real world thereby improving its adaptability and resilience. Notable robotic simulators like MuJoCo (13) and NVIDIA Isaac Sim (14) offer APIs that facilitate domain randomization hence enabling easy manipulation of object properties, physics parameters and the sensor conditions. Moreover, the procedural content generation (PCG) techniques can further automate the creation of diverse environments. By training agents in such varied scenarios, the domain randomization helps to develop more robust control policies that are more likely to succeed in real-world applications even when faced with unexpected situation states and variations.

- **Domain Adaptation**: The domain adaptation is an advanced Sim2Real transfer approach that minimizes the discrepancies between the simulated and real-world domains by aligning their feature distributions accordingly. This method often involves training a model on simulated data and then adapting it to the real-world conditions by either the learning mapping functions or fine-tuning the hyper-parameters of the model under discussion. The approaches such as adversarial training where a generator produces realistic real-world data from simulations while a discriminator distinguishes between them play an important role in this process. Additionally, the feature alignment methods like Maximum Mean Discrepancy (MMD) and the transformation matrices help to establish a common feature space between these two domains. The instance weighting further refines the adaptation by assigning importance to the simulated samples based on their similarity to the real-world data. While the domain adaptation can handle complex domain shifts and often outperforms domain randomization but it typically requires a small set of labelled real-world data for effective guidance and can be computationally intensive particularly within the complex agile automation scenarios. Implementing domain adaptation usually involves deep learning frameworks like PyTorch and TensorFlow which provide comprehensive libraries and tools for adversarial training, feature alignment and instance weighting.

- **Progressive Networks (Curriculum Learning)**: Progressive networks as a Sim2Real transfer technique is related to curriculum learning that focuses upon gradually increasing the complexity of simulation environments to train neural networks more effectively. Initially, the network is trained in simplified settings with basic models or low-resolution perceptual feedback. As training progresses, the complexity of the tasks increases thereby incorporating diverse object shapes, textures, lighting conditions and sensor noise. This incremental approach allows the network to learn hierarchical task representations thereby starting with low-level features in simple environments and building upon them with high-level features as complexity grows. Progressive networks often use a multi-headed architecture where each stage has a separate output head trained on the corresponding level of simulation complexity. These heads may share layers across stages or remain independent with some lateral connections facilitating the knowledge transfer between stages. Robotic simulators like MuJoCo, Gazebo (15), and VREP (16) are commonly used in conjunction with the curriculum learning tools to implement such approach. Although the design and training of progressive networks are complex, this method has shown promise in handling tasks with significant reality gaps thus enabling the network to learn more robust representations that generalize well towards the real-world conditions.

4. Benchmarking Case Studies

- **Bin-Picking Task**:

In this task, a digital twin (**AI-enhanced**) of an industrial-grade robotic cell is developed using the Unity engine. The cell features a KR3 robot equipped with a Schunk EGP50 parallel gripper mounted on a turnkey aluminum frame. For enhanced safety and workspace dependability, a light curtain is installed at the corners of the cell to prevent accidental intrusions. The working objects consist of medical components, including cubes, mushrooms, and knobs, as illustrated in **Figure 4**. To enable the robot to perform optimal grasping and precise placement of these objects, a robust perceptual pipeline for object detection and localization is required. To achieve this, a synthetic dataset was generated using a domain randomization plugin within the Unity engine. This plugin introduces 18 different types of rendering and physical parameter variations, such as lighting, textures, and object positions, in a non-uniform manner to diversify and generalize the data sample distribution effectively. The generated dataset was utilized to train two-stage object localization models, including Faster R-CNN (17), SSD (18), and DOPE (19) models. These models are crucial for accurately detecting and localizing objects within the cluttered environment of the bin, ensuring that the robot can reliably perform grasping and placement tasks. The integration of these models within the digital twin framework allows for the simulation of real-world conditions and the optimization of control policies, ultimately facilitating a more effective Sim2Real transfer for the bin-picking task.

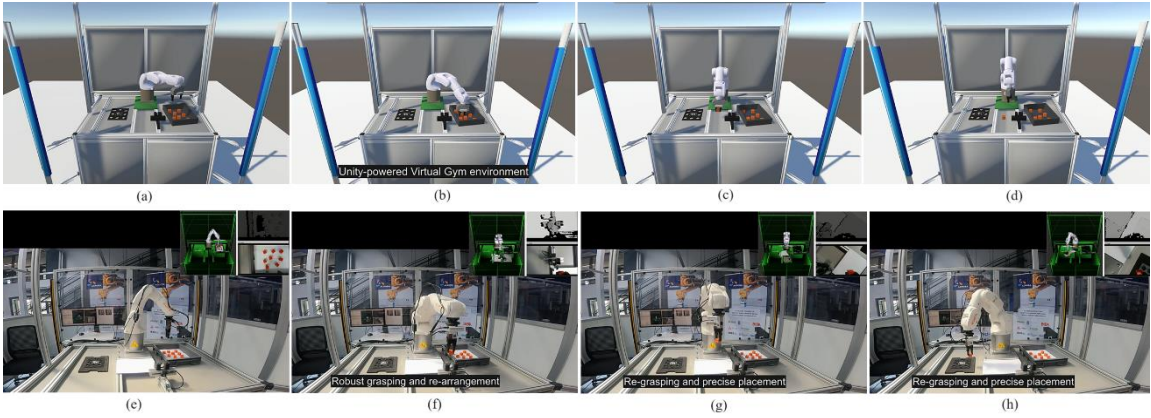

Figure 4. Industrial bin-picking task performed under simulated and real world domain (a-d) show picking, aligning, repositioning, and placement of cubes into chambers within AI-enhanced simulated settings, (e-h)illustrates similar actions over real robotic setup under semi-stochastic conditions.

Among the models evaluated, Faster R-CNN demonstrated superior performance and was subsequently integrated into the perception pipeline for object localization and grasping. In this setup, a camera mounted on the robot's wrist—commonly referred to as an eye-in-hand configuration—first localizes the workspace using ArUco marker detection. Once the robot hovers above the workspace containing candidate objects in a tray, as depicted in **Figure 4**, the perceptual pipeline identifies and localizes the objects in the bin, estimating their 6D poses from the detected bounding boxes using the perspective-to-point (P2P) strategy. This 6D pose information is crucial as it provides precise spatial orientation and positioning data for each object. The 6D pose data is then utilized by the configured MoveIt controller operating in Cartesian space to plan the robot's trajectory, ensuring accurate positioning above the targeted object. The gripper aligns with the object's centroid and closes to grasp it securely. After successfully grasping the object, the robot transports it to the designated placement station, with the MoveIt commander dynamically planning the trajectory for this operation. Upon reaching the placement station, the system localizes the appropriate object holder, and the object is precisely placed into it. This entire process, from perception to motion planning and execution, is seamlessly applied to the real-world setup without the need for retraining or fine-tuning, as shown in **Figure 4**. The robustness and effectiveness of the digital twin backend enable this direct Sim2Real transfer, ensuring minimal to negligible fidelity gaps between the simulation and reality. The system's Sim2Real transfer capabilities were evaluated using KPIs such as repeatability rate, cycle time, reprogramming time, and localization accuracy, as detailed in **Table 1**. The performance variation between the simulated environment and real-world conditions was found to be less than 2%, a significant improvement compared to the higher discrepancies commonly reported in the literature (20)(21). This exceptional performance is largely attributed to the customized randomization plugin developed for this framework, which enhances the robustness, versatility, and agility of the system, making it highly effective in real-world applications.

**Table 1. Performance benchmarking of Industrial Bin-picking task**

| KPI | Digital Twin Environment | Real World Setup |
|---|---|---|
| Repeatability Rate | ~ 5.5 mm precision | ~ 6.8 mm precision |
| Cycle Time | ~ 12 mins, 32 secs | ~ 13 mins, 42 secs |
| Reprogramming Time | ~ 45 mins, 16 secs | ~ 45 mins, 49 secs |
| Localization Accuracy | mAP~ 80.45% | mAP~ 79.55 |

- **Part Inspection:**

In this task, a collaborative robot equipped with a high-precision camera is used to inspect medical parts within both simulated settings (**physics-based approach**) and real-world environments, as shown in **Figure 5**. The setup includes a UR5 collaborative robot arm paired with a micro-level precision scanner, along with a high-performance system featuring a Quadro A4000 GPU to enable faster and parallel processing of inspection scans for both surface and subsurface defect and irregularity analysis. The inspection process begins with a Zivid 3D camera, which captures 360-degree scans of the part under inspection by maneuvering the robot along circular and elliptical trajectory profiles. These scans are then pre-processed to remove any outliers to ensure that the clean and accurate dataset is obtained. The pre-processed data is subsequently fed to the 3D reconstruction algorithm (22) that utilizes the principle of stitching views to build high-fidelity model of object under test. Such reconstructed model exhibits high precision thereby capturing even the minute details that are essential for quality analysis. Once the 3D model is generated then it is compared against its ground truth reference model, which, in this case, is the CAD model of the part. Any discrepancies between the reconstructed and the CAD models are highlighted, extracted and segmented to identify and locate defects on the candidate object. This defect identification process is crucial for post-processing and quality assurance thus ensuring that the final product meets the stringent industry standards.

The performance of the inspection pipeline relies on several critical factors both the internal and external. The internal factors include computational power, camera resolution, calibration accuracy, noise levels and robustness of the 3D reconstruction algorithm. The external factors, such as lighting conditions, surrounding reflections and background variations also play a significant role. In the digital twin environment, such factors can be meticulously controlled which makes Sim2Real transfer evaluation of this task particularly quite challenging. This challenge is further amplified when inspecting industrial parts having complex geometries where even minor discrepancies between the simulated and real-world conditions that can impact the outcome. Upon transferring the inspection pipeline to the real world, the fine-tuning of hyper-parameters at both the camera and algorithmic levels is often necessary to achieve comparable performance as compared to simulated environment. Such a fine-tuning ensures that the system is able to adapt to real-world variations that are not fully captured by the digital twin systm. To validate the system's performance across both domains and assess any discrepancies, KPIs such as resilience score, reconstruction accuracy, inspection time, and defect detection loss are employed, as detailed in **Table 2**. The proposed inspection pipeline consistently achieves a performance level of over 95% in both simulated and real-world scenarios. The remaining 5% discrepancy is primarily due to the slight fidelity gap that current digital twin technologies are unable to completely bridge. This gap underscores the need for further advancements in digital twin technology, particularly in leveraging model surrogacy techniques to enable zero-shot model-to-real transfer performance, thereby eliminating the need for post-transfer hyper-parameter adjustments.

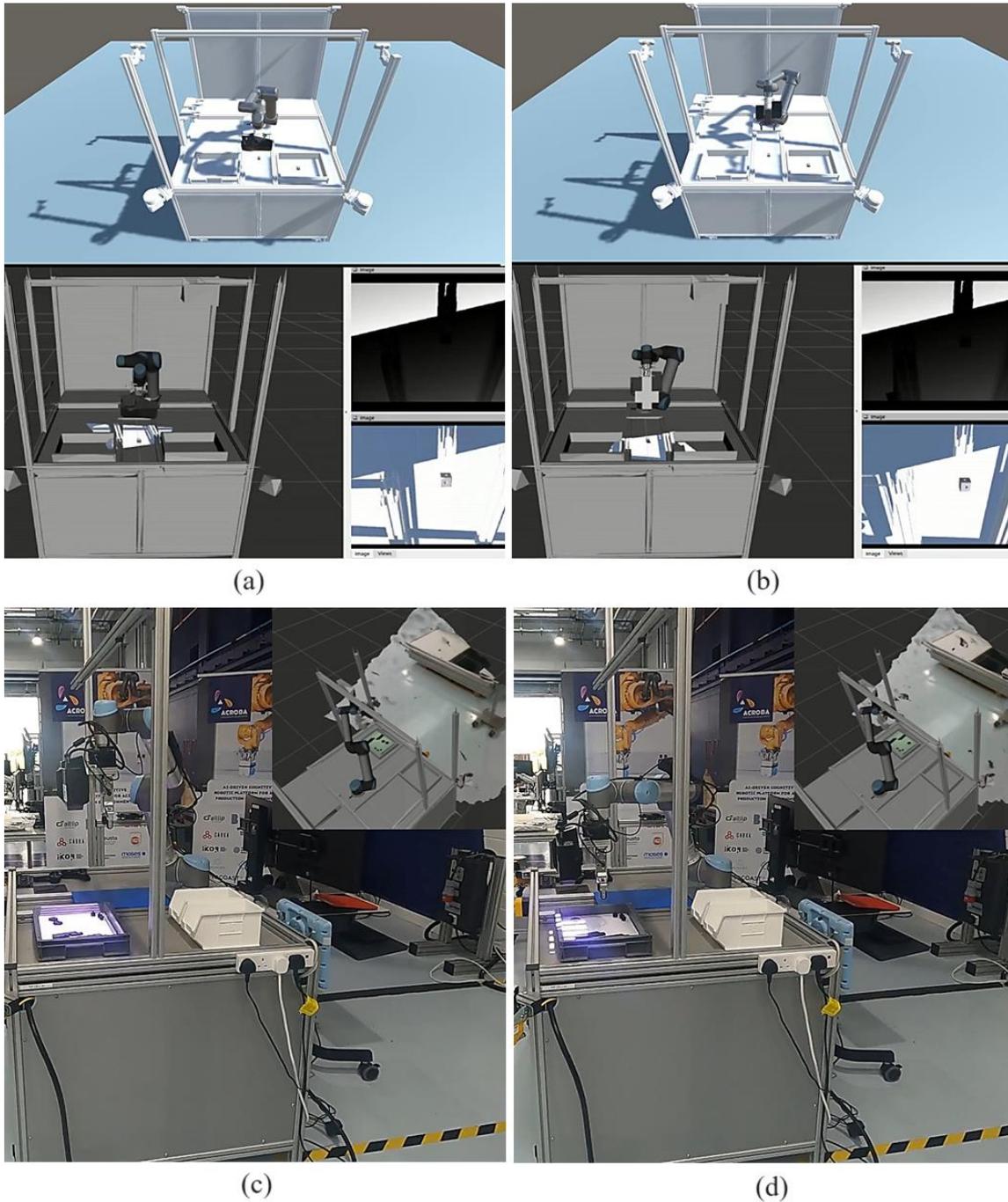

Figure 5. Industrial medical part inspection task, (a-b) represent how 3D reconstruction, CAD compare and defect segmentation are executed in physics-based digital twin, (c-d) show real world demonstration of inspection pipeline using collaborative robotic system.

### Table 2. Performance evaluation of Industrial Inspection task

| KPI | Digital Twin Environment | Real World Setup |
|---|---|---|
| Resilience Score | ~ 0.9234 | ~ 0.9187 |

| Reconstruction Accuracy | SSIM ~ 95.46% | SSIM ~ 91.32% |
|---|---|---|
| Inspection Time | ~ 9 mins, 23 secs | ~ 11 mins, 13 secs |
| Defect Loss | rMSE ~ 3.8% | rMSE ~ 4.3% |

- **Product Assembly**:

In this task, a mobile robot autonomously navigates the workspace, picks parts from production machines, inspects them, and assembles them into the final product: a gearbox. The gearbox assembly consists of several components, including a base plate, cantilever shaft, spur gear wheels with different tooth counts, and nuts for securing the housing. The mobile robot used in this process is the KUKA KMR machine, which features a mecanum wheel-based base for omnidirectional movement, a collaborative LBR iiwa 14 robot arm for flexible manipulation, and an adaptive T-Schunk gripper for precise handling of various parts. The robot is equipped with an eye-in-hand camera mounted on the wrist of the iiwa arm, enabling accurate object localization. The gripper attached to the iiwa arm grasps the identified parts and places them into a tray positioned on the KMR's base. Once all required parts are collected, the mobile robot drives to an inspection station for quality control. Here, the parts are thoroughly checked for possible irregularities or defects using the same inspection pipeline described in the previous use-case. Following successful inspection, the robot transports the parts to the assembly station. At the assembly station, the robot follows task protocols to sequentially assemble the components, constructing the final gearbox prototype. The entire assembly process, from part picking to final assembly, is first performed and validated in a digital twin environment. The trained models and control policies are then seamlessly transferred to the real-world setup, as illustrated in **Figure 6**. This Sim2Real transfer ensures that the robot's performance in the physical environment mirrors the optimized procedures established in the digital twin, thereby reducing the need for extensive post-transfer adjustments.

The digital twin environment (**hybrid approach**) for this setup utilizes a combination of Unity (23) for rendering and MuJoCo for physics modeling. This integration allows for high-quality visualization and accurate physical interactions within the simulated environment. The performance of the digital twin has been assessed using KPIs such as reprogramming time, rework rate, cycle time, and throughput, as detailed in **Table 3**. The results indicate a disparity of over 20% between the KPI values in the simulated environment and those observed in the real world. This significant difference arises from the current limitations in fully simulating production processes within robotic digital twins. The emulation lacks comprehensive modeling of certain design, production, and operational parameters, leading to discrepancies in how the simulated robot perceives and interacts with parts compared to its real-world counterpart. Consequently, the same model and control policy may not guarantee identical performance across both domains. Moreover, the navigation and manipulation capabilities of the mobile robot are affected by the idealization inherent in laser scanner data used for map creation and the imperfect modeling of contact dynamics during manipulation tasks. These factors, influenced by stochastic variations and unmodeled effects, are challenging to replicate accurately in a simulated environment. As a result, the digital twin struggles to fully capture the nuances of real-world interactions, leading to performance deviations. Addressing these disparities requires further advancements in digital twin technologies, particularly in the areas of stochastic modeling and the inclusion of unmodeled physical effects. Enhancing the fidelity of these models will be crucial for reducing the gap between simulation and reality, ultimately leading to more reliable and accurate Sim2Real transfers.

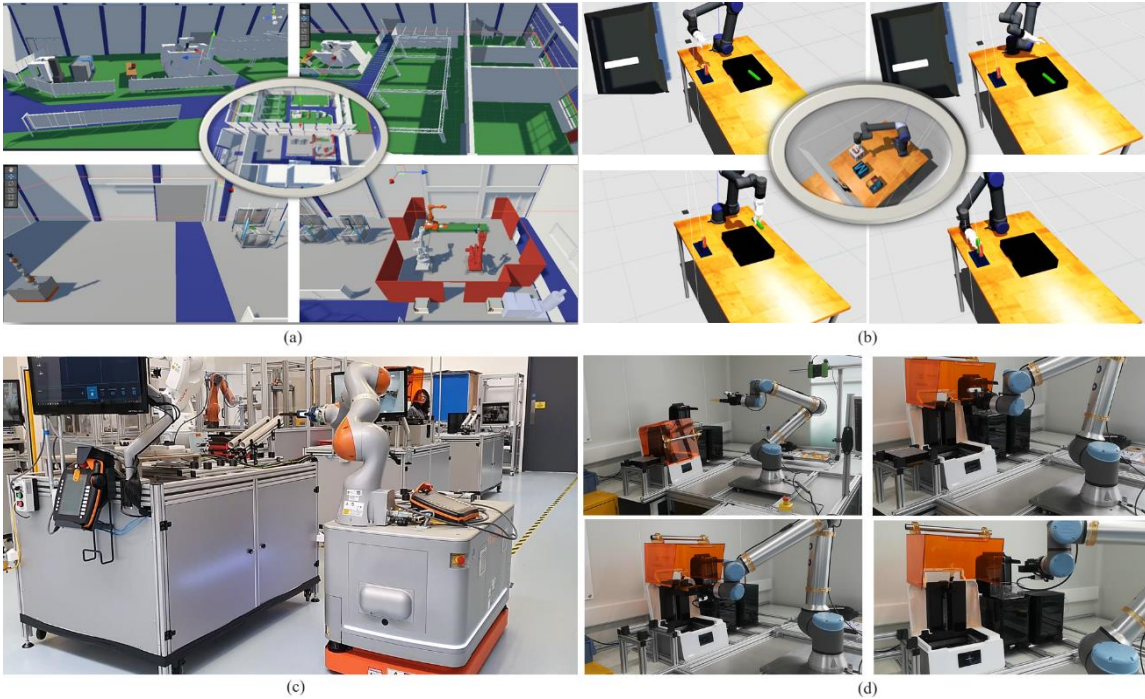

Figure 6. Discrete product assembly of industrial-graded spur gear boxes under dynamic conditions, (a-b) depict navigation, manipulation and environmental interaction within data-driven digital twin settings, (c-d) represent real world implementation and realization of trained models and policies over collaborative mobile agile production line.

**Table 3. Quantitative assessment of Industrial Assembly Process**

| KPI | Digital Twin Environment | Real World Setup |
|---|---|---|
| Reprogramming Time | ~ 32 mins, 52 secs | ~ 37 mins, 14 secs |
| Rework Rate | ~ 0.315 units/batch | ~ 0.385 units/batch |
| Cycle Time | ~ 41 mins, 36 secs | ~ 47 mins, 18 secs |
| Throughput | ~ 7.8 batch units/day | ~ 6.9 batch units/day |

5. Remarks and Recommendations

The application of digital twins and Sim2Real transfer technologies into the agile manufacturing tasks as discussed in this chapter, highlights significant advancements into the high-fidelity simulations, robotics, and machine learning technologies. However, there are still nuances of the real world that remain challenging to fully model and capture especially through multi-sensory feedback and dynamic environment interactions. The scenarios benchmarked in this chapter under both simulated and real domains using practical KPIs unfolds key areas of improvement necessary to scale up these technologies and apply them to new task conditions and domains. They include:

- **Creating Highly Accurate Digital Twin of Robotic Bin-Picking Cells Using Unity Engine:** This involves deploying robust object detection and localization pipeline that is based on synthetic datasets and advanced machine learning models, able to achieve less than 2% Sim2Real transfer variation. To further enhance this process, it is important to establish the real-time feedback loops between the physical setup and the digital twin environments, allowing for continuous updates and refinements of the simulated modeling. For scalability and adaptability to other task conditions, incorporating more sophisticated synchronization and control algorithms are essential to ensure that the digital twin can dynamically adjust to changing conditions.
- **Implementing Collaborative Robotic Systems within Digital Twin for High-Precision Inspection of Medical Parts:** This setup has demonstrated a high-fidelity 3D reconstruction and defect detection capabilities of proposed setup, maintaining over a 95% performance consistency between the simulated and real-world conditions. This highlights the robustness of inspection pipeline. To further improve this process, the integration of real-time defect detection and analysis need to be taken into consideration. Such enhancements would expedite the inspection process, leading to faster iterations, improved throughput and increased productivity within the agile manufacturing settings.
- **Enhancing the KUKA KMR Machine's Navigation, Inspection and Assembly Capabilities using Hybrid Digital Twin Settings:** The current performance under the real-world settings shows a 20% discrepancy compared to its simulated counterpart. To close such domain gap, improving the fidelity of hybrid digital twins by incorporating more detailed models of physical interactions and dynamics is desired. This includes better simulation of the contact forces, the material properties and the environmental variability. Additionally, employing adaptive learning strategies that allows the mobile robot to adjust its navigation and manipulation plans based on the real-time sensory feedback is key to reducing the performance differences between the simulated and real-world settings.

Considering the strengths and limitations of all the discussed digital twin approaches and Sim2Real transfer technologies exercised within the three different use cases considered in this chapter, the following improvements and future directions are proposed to advance the fields. These enhancements are designed to maximize the productivity, throughput and resilience within the agile manufacturing industry through more flexible and efficient robotics approaches.

- **Advanced Simulation Environments:** Current robotic simulators face challenges that restricts their effective Sim2Real transfer, chiefly due to limitations in their high-fidelity physics simulations. These simulations often struggle to accurately capture the material properties, deformable objects and intricate robot-environment interactions, leading to a marginal gap between the simulated and real-world performances. Addressing this gap requires improved sensor realism, that is achieved by incorporating noise, distortion and other artifacts into the simulated sensory data to better reflect real-world imperfections. Moreover, implementing closed-loop simulations that integrate real-time data from physical sensors into the simulated environment is crucial and important. This integration allows the simulation to dynamically adapt to changing conditions thereby supporting more accurate and reliable feedback for active control systems. Collectively such advancements would enhance the realism and predictive accuracy of simulations and hence support the development of more robust and flexible control policies for real-world deployment (24).

- **Generative Data-driven Modeling:** Existing data-driven approaches for Sim2Real transfer often rely upon large synthetic data, which can imped the robustness of the resulting models. A more effective approach involves incorporating real-world data through a real2sim2real pipeline that captures sensor readings, actions and outcomes directly from the physical environment to reflect the real-world nuances. By leveraging unsupervised and semi-supervised learning techniques such as Generative Adversarial Networks (GANs) can enable the generation of near-real data and hence reduce dependence over the expensive labeled datasets. Additionally, the transfer learning from pre-trained models that have been fine-tuned with a smaller real-world dataset, can accelerate the learning process and enhances the Sim2Real transfer performance. Continuous improvement of control policies as new data becomes available ensures that the robotic systems remain agile, efficient and adaptable to stochastic conditions.

- **Hybrid Differential Twins:** Currently, no commercial differential digital twin solution is available at a scale that is viable for industrial automation research and prototyping. To address such gap, developing a hybrid system that combines the high-fidelity rendering with accurate physics modeling is essential. This system should be differential, that means it should be capable of modeling highly dynamic motion in the articulated systems with both high accuracy and computational efficiency. Key features should include; implicit numerical integration, fractional contact handling and gradient descent approaches for mapping real-world robot motions and environmental feedback to the control inputs within the simulated settings. This hybrid system that is envisioned as a blend of technologies like PyBullet, MuJoCo, and Isaac Gym would enhance the fidelity of digital twins by dynamically adjusting the simulation parameters based on the real-world performance . Additionally, it would leverage online adaptation techniques, such as Bayesian optimization, to continuously improve the simulation accuracy and ensures optimal performance across varying task conditions in agile manufacturing settings (25).

- **Cloud Robotics Automation:** The deployment and service of existing hardware, including GPUs, TPUs and the various sensors (e.g., vision, tactile, LiDAR, ultrasonic) accounts for significant costs and require extensive local integration and support. The complexity and high resource demands limit widespread adoption, particularly within industries that are lacking in-house expertise. To overcome such challenges, the development of cloud-based multi-modal robotics services is highly recommended. These services would provide digital twins the integration of APIs and the support for physics-based simulations, the complex renderings and the AI training, that all being hosted in the cloud. By offloading computational tasks to cloud resources, this approach would ensure scalability, modularity and collaborative capabilities. It would also enable seamless communication between the digital and real-world domains through secure

- **Task-specific Optimization:** Current digital twins are generally designed for broad applicability to various domains that cause suboptimal performance under desired bespoke task conditions. This lack of specialization underscores the need for developing more task-specific libraries that are tailored to distinct applications including; manipulation, navigation, human-robot collaboration and the robot learning. For example, in assembly tasks, more precise contact models by using either penalty-based, constraint-based, or compliant methods should be combined with advanced control algorithms (e.g., impedance or force-based control) and error recovery strategies to ensure optimal performance. In bin-picking tasks, the focus should be on refining the object recognition and pose estimation algorithms, as demonstrated by NVIDIA's DOPE within the simulations that train models on synthetic datasets with known ground truth poses to enhance the real-world performance. For inspection tasks, the emphasis should be placed on simulating surface defects, material properties and lighting conditions using surrogate models within physics-based rendering combined with the generative models. This approach would ensure an accurate defect representation, realistic simulation of material properties and the effective modeling of inspection sensor characteristics hence leading to high-quality simulated datasets. By developing and integrating such task-specific libraries, more robust, resilient and reliable automation platforms can be created, enabling seamless deployment in the real-world applications (26).

protocols, thus improving the robotic perception, simulation speed and Sim2Real transfer efficiency. The integration of sensor fusion, the high-performance computing (HPC) and the cloud robotics promises to facilitate more effective and accessible robotic automation deployments across varying domains.

**Appendix:**

This section provides description of KPIs that are used in the three scenarios for performance assessment:

- **Repeatability Rate**: This KPI measures the consistency with which a robotic system can perform the same task many times under homogenous conditions. It is quantified by evaluating the variance in the pose of robot end-effector or/and trajectory during repeated executions. A higher value of repeatability rate is crucial for tasks requiring precision, such as assembly and inspection where even some minor deviations can affect the overall quality of the end product.
- **Cycle Time**: Cycle time refers to the total time required to complete a single iteration of a task from start to goal. This KPI is critical for assessing the efficiency of manufacturing processes as it directly affects the overall production rate. In the realm of agile manufacturing, minimizing cycle time without hampering quality is of utmost importance for optimizing the throughput and satisfying the dynamic production demands.
- **Reprogramming Time**: The reprogramming time measures the duration that is required to modify or update the control logic of a robotic system in response to changes in the task requirements or/and environmental conditions. This KPI is essentially important for agile manufacturing settings, where rapid adaptation to new tasks or/and products is necessary. The lower value of reprogramming times indicates greater system flexibility and adaptability.

- **Localization Accuracy**: This KPI measures the precision with which a robotic system can determine the 6D pose of candidate objects within its workspace. The high value of localization accuracy is pivotal for tasks such as object manipulation, part assembly and inspection, where the precise positioning is required to ensure successful task execution. The localization accuracy is typically evaluated by comparing the pose detected object with its ground truth label.
- **Resilience Score**: The resilience score evaluates the ability of robotic system to maintain its performance under disturbances, uncertainties and variations during operating conditions. This KPI is a measure of the system's robustness that reflects its capacity to recover from errors and adapt to changes without any significant compromise into the performance.
- **Reconstruction Accuracy**: The reconstruction accuracy refers to the fidelity with which a 3D model is created by the reconstruction algorithm that represents the actual physical object. This KPI is useful for tasks that involve the inspection and quality control of objects, therefore the precise modeling of objects is necessary to detect the defects or/and deviations from the ground-truth data. The higher reconstruction accuracy leads to more reliable inspections and better quality assurance accordingly.
- **Inspection Time**: The inspection time measures the duration that is required for a robotic system to complete the inspection of an object or/and assembly under test. This KPI is important for assessing the efficiency of quality control processes where the timely identification of defects is needed to prevent the bottlenecks and ensures high throughput in the existing production lines.
- **Defect Loss**: The defect loss quantifies the proportion of products or/and the components that fail to meet the quality standards during the inspection process and must be either reworked or discarded. This KPI is a direct measure of effectiveness of the inspection process and the overall quality of the agile manufacturing output. The lower defect loss indicates higher manufacturing quality and efficiency.
- **Rework Rate**: The rework rate represents the percentage of products that require additional processing or correction after initial production due to defects or errors identified during the inspection of them. This KPI is a key indicator of production quality and process reliability. The lower rework rates indicate a more efficient and accurate agile manufacturing processes that ensure reduced waste and production costs accordingly.
- **Throughput**: The throughput is defined as the rate at which a manufacturing system produces parts and products within given time frame. This KPI is critical for evaluating the productivity of agile manufacturing processes especially for high-demand production setups. The high value of throughput with low defect rates and rework indicates well-optimized production ecosystem that is capable of meeting the market demands more efficiently and effectively.

Acknowledgement

This work is supported by EU funded CORESENSE and ACROBA projects under grant agreement No 101070254 and agreement No 101017284. Authors are grateful to IMR, Ireland for providing research oriented environment and opportunity to undertake this study with a view towards the existing manufacturing industry outlook.